\documentclass[11pt,a4paper]{article}
\usepackage{authblk}
\usepackage[hyperref]{acl2020}
\usepackage{times}
\usepackage{latexsym}

\usepackage{multirow}
\usepackage{graphicx}
\usepackage{amsmath,amssymb}
\usepackage{siunitx}
\usepackage{dblfloatfix}

\usepackage{microtype}

\aclfinalcopy 
\def\aclpaperid{1319} 

\newcommand{\conpono}{{\sc Conpono}}

\title{Pretraining with Contrastive Sentence Objectives Improves Discourse Performance of Language Models}

\author[*\thanks{\quad Work done during internship at Google.}]{\textbf{Dan Iter}}
\author[**]{\textbf{Kelvin Guu}}
\author[**]{\textbf{Larry Lansing}}
\author[*]{\textbf{Dan Jurafsky}}
\affil[*]{Computer Science Department, Stanford University}
\affil[**]{Google Research}
\affil[ ]{\texttt {\{daniter,jurafsky\}@stanford.edu}}
\affil[ ]{\texttt {\{kguu,llansing\}@google.com}}

\date{April 2020}

\begin{document}
\maketitle
\begin{abstract}\vspace*{-1ex}
Recent models for unsupervised representation learning of text have employed a number of techniques to improve contextual word representations but have put little focus on discourse-level representations.
We propose \conpono{}\footnote{Code is available at https://github.com/google-research/language/tree/master/language/conpono and https://github.com/daniter-cu/DiscoEval}, an inter-sentence objective for pretraining language models that models discourse coherence and the distance between sentences.
Given an anchor sentence, our model is trained to predict the text $k$ sentences away using a sampled-softmax objective where the candidates consist of neighboring sentences and sentences randomly sampled from the corpus.
On the discourse representation benchmark DiscoEval, our model improves over the previous state-of-the-art by up to 13\% and on average 4\% absolute across 7 tasks.
Our model is the same size as BERT-Base, but outperforms the much larger BERT-Large model and other more recent approaches that incorporate discourse.
We also show that \conpono{} yields gains of 2\%-6\% absolute even for tasks that do not explicitly evaluate discourse: textual entailment (RTE), common sense reasoning (COPA) and reading comprehension (ReCoRD). 
\end{abstract}

\section{Introduction}



Pretraining large language models has become the primary method for learning representations from unsupervised text corpora.
Since the initial improvements demonstrated by ELMo \cite{peters-etal-2018-deep} and BERT \cite{devlin-etal-2019-bert}, many alternative pretraining methods have been proposed to best leverage unlabeled data.
These methods include bi-directional language modeling \cite{peters-etal-2018-deep}, masked language models \cite{devlin-etal-2019-bert}, word order permutation \cite{yang2019xlnet}, more robust training \cite{liu2019roberta} and more efficient architectures \cite{lan2019albert}.
However, little focus has been put on learning discourse coherence as part of the pretraining objective.

While discourse coherence has been of great interest in recent natural language processing literature \cite{chen-etal-2019-evaluation, nie-etal-2019-dissent, xu-etal-2019-cross}, its benefits have been questioned for pretrained language models, some even opting to remove any sentence ordering objective \cite{liu2019roberta}.
However, in a recently published benchmark for evaluating discourse representations, \citet{chen-etal-2019-evaluation} found that the best performing model was surprisingly BERT, despite comparing against models specifically designed for discourse, such as DisSent \cite{nie-etal-2019-dissent} and a new recurrent network trained on a large range of sentence ordering objectives.
We show that combining transformer encoders with our  inter-sentence coherence objective, we can further improve discourse-level representations in language models.


We present a model that trains a sentence-level encoder to capture discourse relationships between sentences, including ordering, distance and coherence.
The encoder is trained by using its output to predict spans of text that are some $k$ sentences away from a context in either direction.
The predictions are made discriminatively with a sampled-softmax that contrasts the correct target sentence against negatives, including hard examples sampled from the same paragraph.
Our objective is inspired by the recently proposed Constrastive Predictive Coding (CPC) \cite{oord2018representation}, but, among other differences, is applied on the sentence-level rather than the token-level and is bi-directional.
We call this the CONtrastive Position and Ordering with Negatives Objective (\conpono{})\footnote{Also means \textit{arrange} or \textit{order} in Latin.}.



We evaluate our model on DiscoEval \cite{chen-etal-2019-evaluation}, a recently published benchmark for evaluating and probing for various aspects of discourse-level semantics in representations output by discourse models.
We observe that the representations learned with \conpono{} outperform BERT-Large and achieve a new state-of-the-art despite using fewer parameters and training on the same data.
Furthermore, we show that our new objective improves model performance on other tasks including textual entailment, common-sense reasoning and reading comprehension.
We compare \conpono{} against BERT-Base on RTE \cite{giampiccolo2007third,bentivogli2009fifth}, COPA \cite{roemmele2011choice} and ReCoRD \cite{zhang2018record}, while controlling for model size, training data and training time.


Our main contributions are:

    
    

\begin{enumerate}
    \item  We describe a novel sentence-level discourse objective that is used in conjunction with a masked language model for unsupervised representation learning for text. We show that this objective can leverage the cross-attention and pretrained weights of a transformer model to learn discourse-level representations.
    \item We show that our model achieves a new state-of-the-art on DiscoEval, improving the results on 5 of the 7 tasks and increasing accuracy by up to 13\% and an average of over 4\% absolute across all tasks. We also show 2\%-6\% absolute improvements over Bert-Base on RTE, COPA and ReCoRD as evidence that discourse pretraining can also improve model performance on textual entailment, commonsense reasoning and reading comprehension.
\end{enumerate}

\section{Model}\label{sec:model}

\begin{figure*}[h]
\includegraphics[width=16cm]{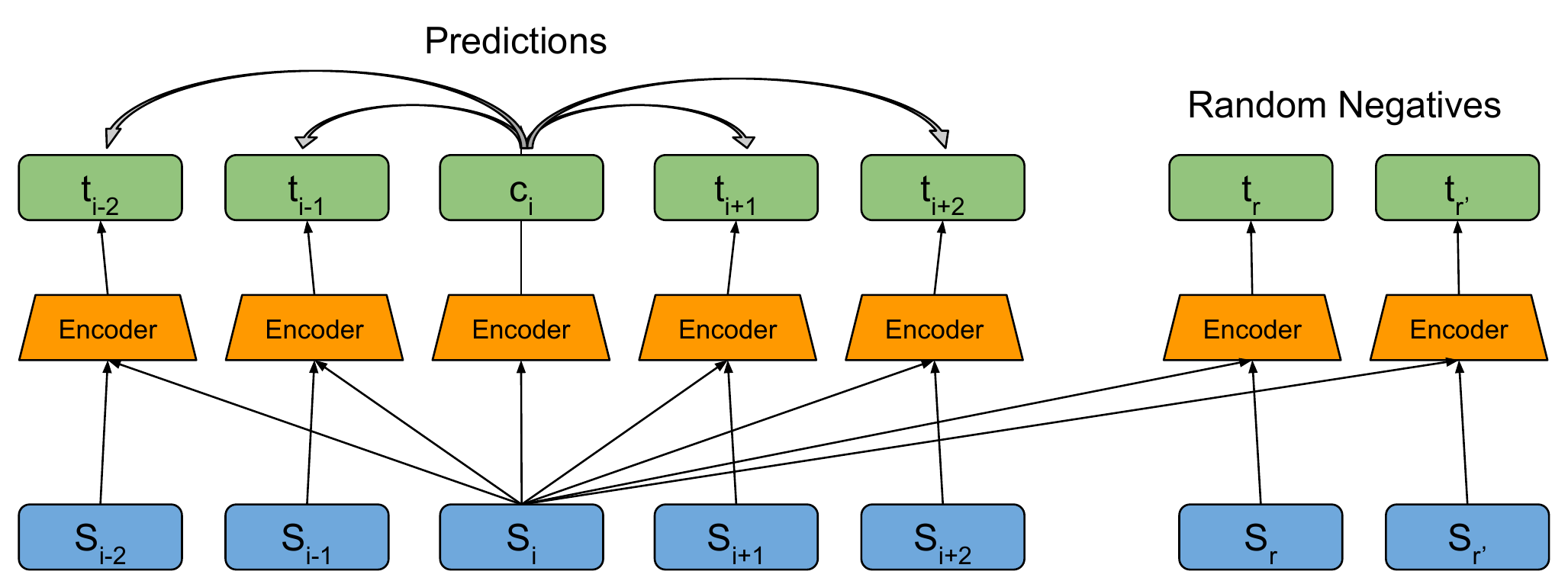}
\caption{During training, a text segment is selected as the anchor ($S_i$).
The anchor as well as all the targets, $S_{i-2}...S_{i+2}$ plus random samples $S_{r}$ are encoded with the transformer masked language model.
The encoded representation of the anchor is used to predict each target at its target distance. The $S_i$ objects are raw text sentences, the \textit{encoder} is the transformer model, and $c_i$ and $t_i$ are vectors.}
\label{fig:conpono}
\end{figure*}

Figure~\ref{fig:conpono} illustrates the \conpono{} model.
The intuition is that if the model is able to accurately predict the surrounding target sentences given some anchor text, then the vector representations for these sentences should also be useful for downstream tasks.

The input to the model is a paragraph that is split into sentences.
A sentence is chosen at random as the anchor, and will be denoted as $s_i$.
We encode $s_i$ with a transformer encoder to produce a vector $c_i$.
The surrounding sentences are denoted as $s_{i+k}$ where $k \in [{-K}\:..\:{-1},1\:..\:K]$, meaning the maximum distance we use is $K$.
We report results for $K \in [1..4]$.
These sentences, $s_{i+k}$, are encoded jointly with the anchor sentence.
We use just a single encoder $g_\theta$ so all text is encoded with the same weights.
The encoded vectors are named $t_{i+k}$ because these are the target vectors the model tries to identify given the anchor and a target distance $k$.
Equation~\ref{eq:def} defines $t_{i+k}$ and $c_i$ as a function $g_\theta$ of the input sentences.
Note that the \conpono{} $g_\theta$ is different from the encoder in CPC because we input both the anchor and the target into the encoder, rather than separate anchor and target encoders.

\begin{equation}
    t_{i+k} = g_\theta (s_i, s_{i+k}),\quad c_i = g_\theta(s_i)
    \label{eq:def}
\end{equation}

Given the anchor and targets, we define a log-bilinear model in equation~\ref{eq:cpc} to score the plausibility of target $t_{i+k}$ being in position $k$ from anchor $c_i$.
The full set of parameters for our model is $\theta$ for the encoder and a $W_k$ for each $k$.
CPC has the same bi-linear form as Equation~\ref{eq:cpc} but the architecture for the encoders is different.

\begin{equation}
    f_k(s_{i+k}, s_i) = \exp(t_{i+k}^T W_k c_i)
    \label{eq:cpc}
\end{equation}

The loss for each $k$ is given in equation~\ref{eq:loss} where the score for the correct target is contrasted to scores of random samples $s_j$, sampled from both in-document and random sentences from the corpus, $S$.

\begin{equation}
    \mathcal{L}_k = -\mathbb{E}_S \left[ log 
    \frac{f_k(s_{i+k}, s_i)}{\Sigma_{s_j \in S} \; f_k(s_j, s_i)} \right]
    \label{eq:loss}
\end{equation}

To train \conpono{}, we sample negative examples randomly from the corpus and from the same paragraph but different $k$ as hard negatives.
Note that when $|k|$ is greater than 1, there will be sentences between the anchor sentence and target sentence that will be purposely omitted from the input.
The missing context is intended to create a challenging objective where the model may not be able to rely on trivial signals that often appear in contiguous sentences. 

\subsection{Encoder Architectures}\label{sec:encoding}

For each example we encode two text spans, the anchor and the target.
There are three main options for encoding the two spans into $c_i$ and $t_{i+k}$.
The simplest method, and most similar to CPC is to encode the anchor and target separately, which we call \textit{isolated encoding}. 
With this encoder, equation \ref{eq:def} will be $t_{i+k} = g_\theta (s_{i+k})$.
The major drawback of this approach is that there is no token-level cross-attention between the anchor and the target, which has been shown to generally improve text encoding \cite{vaswani2017attention}.
Cross-attention is the mechanism in neural networks that allows for attention to be shared between multiple inputs, in our case, two separate spans of text.

Alternatively, we can encode the anchor and target together and then dot product the latent vector with a learned vector representation for each distance $k$.
We call this approach a \textit{uni-encoder}.
With this encoder, equation \ref{eq:cpc} will be $f_k(s_{i+k}, s_i) = \exp(t_{i+k}^T w_k) $.
The class matrix $W_k$ in equation \ref{eq:cpc} is replaced by a class vector $w_k$, which has fewer parameters.
This is similar to the ordering objectives in BERT and ALBERT where the pooled representation is used for a binary classification task and the learned vector representation for each distance $k$ is just the softmax weights.
The potential drawback to this method is that each pair of sentences is represented by a single vector.
This encoder may learn a representation that is similar for all examples that have the same label but does not explicitly model the content of the input.

\conpono{} implements the intersection of these two approaches.
The targets are concatenated to the anchor when encoded, to make use of the cross-attention of the transformer encoder.
The anchor, is encoded independently, though with the same weights.
This objective allows for more freedom in the values of $c_i$ and $t_{i+k}$, unlike the \textit{uni-encoder}.
Furthermore, since the encoder, $g_\theta$, can encode either one span ($s_i$) or two spans ($s_i, s_{i+k}$), it can be used for downstream tasks that have either single (eg. SSP) or double (eg. BSO) span inputs.

\subsection{Comparing Inter-Sentence Modeling Objectives}\label{sec:sent_objectives}

There are different \textit{tasks} that can be used for learning inter-sentence representations.
BERT \cite{devlin-etal-2019-bert} included a next sentence prediction (NSP) task.
For NSP, two spans are fed into the model with the second span either being the next contiguous span of text from the source or 50\% of the time it is replaced with a random span from the corpus.
The task is a binary classification of whether the two spans are from the same source.
ALBERT \cite{lan2019albert} compares the NSP approach to using no inter-sentence objective and to sentence order prediction, which for clarity we refer to as binary sentence ordering (BSO).
For BSO, the input is two spans that are always contiguous and from the same source but 50\% of the time are in reverse order.
With \conpono{} we capture the benefits of both learning ordering between coherent sentences and contrasting against random negatives.
We make the objective even more challenging by also predicting order on spans that are multiple sentences apart, and using other sentences from the same paragraph as harder negatives.

\subsection{Technical details}

In practice, we use a 512 token input which is much larger than most two sentence pairs.
To train on longer sequence lengths, we use 4 sentences as the anchor and 3 sentences as the target segment.
We truncate longer sentences and pad tokens up to the sequence length as done for typical BERT input.
There is no overlap between the two segments and the $k$ distance refers to the number of sentences omitted between the two segments.
For example, for a paragraph we may choose $s_7..s_{10}$ as the anchor and $s_1..s_3$ as the target for $k=-4$ because $s_3$ is 4 positions behind $s_7$.
Since most paragraphs are not long enough to have many sentences in both directions of a 4 sentence anchor, we randomly select 4 of the 8 possible $k$ targets for a given paragraph.
Because of the random sampling, we oversample shorter distances because they occur more consistently in the data.

We train with 32 input sentences, where 1 is the correct target, 3 are hard negatives from the same document and 28 are random sentences from other documents.
For fair comparison, we train on the same data as BERT, using only Wikipedia and BooksCorpus \cite{zhu2015aligning}.
We initialize our model with BERT-Base weights and train until the model has seen one-fourth as many segment pairs as the original BERT model (~32M total), so the total compute and iterations of training are not significantly greater than BERT-Base.
We also use a masked language model objective similar to BERT but dynamically mask during training for different masks each epoch.
When jointly encoding two inputs, we concatenate the input tokens and separate the two spans with a ``[SEP]'' token to mimic the BERT format.

\section{Evaluation}

We evaluate our model on the DiscoEval benchmark \cite{chen-etal-2019-evaluation} and on the RTE \cite{giampiccolo2007third,bentivogli2009fifth}, COPA \cite{roemmele2011choice} and ReCoRD \cite{zhang2018record} datasets.
We chose the DiscoEval benchmark because it is intended to evaluate a model's ability to represent the ``role of a sentence in its discourse context''.
We also report results on RTE, COPA and ReCoRD because these tasks have a discourse or sentence ordering aspect to them but are not exclusively designed for discourse evaluation.

\subsection{Discourse Evaluation} \label{sec:discourse}

\begin{table*}[h!]
\centering
\begin{tabular}{||c | c c c c c c c c||} 
 \hline
 Model & SP & BSO & DC & SSP & PDTB-E & PDTB-I & RST-DT & avg. \\ 
 \hline\hline
 BERT-Base & 53.1 & 68.5 & 58.9 & 80.3 & 41.9 & 42.4 & 58.8 & 57.7 \\
 BERT-Large & 53.8 & 69.3 & 59.6 & \textbf{80.4} & \textbf{44.3} & 43.6 & 59.1 & 58.6  \\ 
 \hline
  RoBERTa-Base & 38.7 & 58.7 & 58.4 & 79.7 & 39.4 & 40.6 & 44.1 & 51.4  \\
  BERT-Base BSO & 53.7 & 72.0 & 71.9 & 80.0 & 42.7 & 40.5 & \textbf{63.8} & 60.6  \\
  \hline
  \conpono{} \textit{isolated} &  50.2  &  57.9  & 63.2 & 79.9 & 35.8 & 39.6 & 48.7 & 53.6 \\ 
  \conpono{} \textit{uni-encoder} & 59.9 & 74.6 & 72.0 & 79.6 & 40.0 & 43.9 & 61.9 & 61.7  \\
 \hline
  \conpono{} (k=2) & \textbf{60.7} & \textbf{76.8} & \textbf{72.9 } & \textbf{80.4}  & 42.9 & \textbf{44.9} & 63.1  & \textbf{63.0}  \\
\conpono{} std. & $\pm .3$ & $\pm .1$ & $\pm .3$ & $\pm .1$ & $\pm .7$ & $\pm .6$ & $\pm .2$ & - \\
 \hline
\end{tabular}
\caption{\conpono{} improves the previous state-of-the-art on four DiscoEval tasks. The average accuracy across all tasks is also a new state-of-the-art, despite a small drop in accuracy for PDTB-E. BERT-Base and BERT-Large numbers are reported from \citet{chen-etal-2019-evaluation}, while the rest were collected for this paper.
We report standard deviations by running the evaluations 10 times with different seeds for the same \conpono{} model weights.}
\label{table:discoeval}
\end{table*}

\textbf{Tasks:} DiscoEval \cite{chen-etal-2019-evaluation} is a suite of tasks 
``designed to evaluate discourse-related knowledge
in pretrained sentence representations''.
The benchmark is composed of seven tasks; four based on sentence ordering or coherence (Sentence position (SP), Binary sentence ordering (BSO), Discource coherence (DC) and Sentence section prediction (SSP)) and three that are based on classifying the type of relationship between a pair of text sequences (Penn Discourse Tree Bank Explicit and Implicit (PDTB-E/I) and Rhetorical structure theory (RST)).
PDTB \cite{prasad-etal-2008-penn} and RST \cite{carlson-etal-2001-building} are human annotated datasets.
Both are multi-class classification tasks where PDTB is classifying a pair of sentences whereas RST is predicting the class of a node in a document-level discourse tree.
Both classes of tasks are critical aspects of understanding discourse.

\textbf{Baselines:}
The previously best overall performing model from DiscoEval \cite{chen-etal-2019-evaluation} was BERT-Large \cite{devlin-etal-2019-bert}.
We also include the results for BERT-Base because our model is most comparable to BERT-Base in terms of parameter size, training data and training compute.
We also evaluate RoBERTa-Base \cite{liu2019roberta} because it was trained on more data, reported improvements over BERT-Base on other tasks but dropped the next sentence prediction objective entirely.
We also compare against a BERT-Base model which we trained with binary sentence ordering (BERT-Base BSO) because this objective has been shown to be more useful than next sentence prediction \cite{lan2019albert}.
This BERT-Base BSO model was initialized with BERT weights and trained on the same data but only on contiguous spans of text where 50\% of the time we switch the order.
This model and \conpono{} are initialized from the same weights and trained on the same number of segment pairs so that the two models can be compared fairly.

In Section~\ref{sec:encoding} we describe different encoding approaches for generating the sentence-level representations.
We report results from versions of \conpono{} using each of these encoding approaches, labeled \textit{isolated} to represent separate encoding and \textit{uni-encoder} to represent joint encoding of the anchor and target without a separate anchor encoding.
The final line in Table~\ref{table:discoeval} is the combined approach that we describe in Section~\ref{sec:model}.

\textbf{Modeling DiscoEval}
We reuse the code from DiscoEval and generally maintain the same process for collecting our results on the benchmark, such as freezing all weights and only training a logistic regression or one layer perceptron on top of the sentence encodings.
Note that since we are only interested in the vector representations of the input, we drop the weight matrix $W_k$ and only use the output of the encoder.
We omit the details for the encoding logic for each task since that is explained in detail in \citet{chen-etal-2019-evaluation}.
Here we only mention our deviations from the \citet{chen-etal-2019-evaluation} methodology.
The most salient difference is that we only use the \textit{pooled} representation from our model rather than the average from multiple layers of the model for the SP, BSO and DC tasks.

For encoding individual tasks we prefer to encode pairs of sentences together.
For SP we encode the first sentence concatenated with every other sentence instead of taking the point-wise difference and concatenate the 5 vectors.
For BSO we also encode the two sentences together instead of separately.
For DC we split the paragraph into pairs of sentences and encode those together. 
We concatenate the 3 output vectors.
For RST instead of embedding each sentence and doing a mean of all the sentences in a subtree, we simply concatenate those sentences and encode them all together as a single text span.
Any text segments longer than 512 tokens are truncated from the end.

\textbf{Results:}
Table~\ref{table:discoeval} shows that our model outperforms the previous state-of-the-art accuracy on DiscoEval overall.
Our model excels in particular on the sentence ordering and coherence tasks (SP, BSO, and DC).
Note that our model parameter count is the same as BERT-Base but it outperforms BERT-Large, which has significantly more parameters and has used much more compute for pretraining.
From the discussion in Section~\ref{sec:sent_objectives}, BERT represents using the NSP objective and we train BERT-Base BSO to compare NSP, BSO and \conpono{} directly.
BERT-Base BSO scores tend to fall between those of BERT-Base and our model, implying that the sentence ordering objective is improving the models for this benchmark, but that binary sentence ordering is not sufficient to capture the added benefits of including more fine-grained ordering and negative examples.

We observe that \conpono{} outperforms both the \textit{isolated encoding} and \textit{uni-encoding} approaches.
\conpono{} \textit{isolated} preforms significantly worse than both other approaches, suggesting that cross-attention between the anchor and the target is critical to learning stronger discourse representations.
\conpono{} \textit{uni-encoder} results are closer to our combined encoding approach but still fall short on every task.
This empirical result suggests that the separate encoding of the anchor during pretraining is important despite the fact that theoretically \conpono{} could trivially reduce to the \textit{uni-coder} representation by ignoring $c_i$.

\subsection{RTE, COPA and ReCoRD}

\begin{table*}[h!]
\centering
\small
\begin{tabular}{p{7.5cm} | p{7.5cm} } 
 \hline
 Context & Completions \\
 \hline
 \multicolumn{2}{ c }{ReCoRD} \\
 \hline
... Despite its buzz, the odds are stacked against \textit{Google}'s \textit{Chrome OS} becoming a serious rival to \textit{Windows}... \textit{Chrome OS} must face the same challenges as \textit{Linux}: compatibility and unfamiliarity.
A big stumbling block for  \textit{Google} will be whether its system supports \textit{iTunes}. & Google will also be under pressure to ensure \underline{[\textbf{Chrome OS} / iTunes / Linux]} works flawlessly with gadgets such as cameras, printers, smartphones and e-book readers. \\ 
 \hline
  \multicolumn{2}{ c }{RTE} \\
\hline
Rabies virus infects the central nervous system, causing encephalopathy and ultimately death. Early symptoms of rabies in humans are nonspecific, consisting of fever, headache, and general malaise. & \textbf{Rabies is fatal in humans.} \\ 
\hline
  \multicolumn{2}{ c }{COPA} \\
  \hline
  The women met for coffee. & \textbf{They wanted to catch up with each other.} \\ [5px]
  & The cafe reopened in a new location. \\ 
\hline
\end{tabular}
\caption{These are examples from ReCoRD, RTE, and COPA that exhibit aspects of discourse coherence. For ReCoRD, candidate entities are in italics and replaced terms in the completion are underlined. True completions are bold.}
\label{table:examples}
\end{table*}

\textbf{Tasks:}
DiscoEval was specifically designed to evaluate model performance on discourse tasks but there are many other benchmarks that could also benefit from pretraining for improved discourse coherence.
We evaluate our model on three such tasks, Recognizing Textual Entailment (RTE) \cite{giampiccolo2007third,bentivogli2009fifth}, Corpus of Plausible Alternatives (COPA) \cite{roemmele2011choice} and Reading Comprehension with Commonsense Reasoning Dataset (ReCoRD) \cite{zhang2018record}.
We report accuracy on the validation set provided by each dataset.

Each example in RTE is a pair of sentences. 
The model must classify whether or not the second sentence entails the first.
Examples in COPA are composed of a single context sentence followed by two candidate sentences that are either a cause or effect of the context sentence. 
The model must select the most ``plausible'' sentence of the two.
Lastly, an example in ReCoRD is a paragraph from a news article, followed by several bullet points and with all the entities marked.
The model is given a single sentence from later in the document with a single entity masked out and must select the entity from the context that fills the blank.
Table~\ref{table:examples} shows examples of each with correct choices in bold.


\textbf{Baselines:}
We compare our model against BERT-Base because this is the closest model in terms of parameter size and training data.
However, since our model is initialized with BERT-Base weights, we also report results from BERT-Base BSO because it was trained on the same number of text examples as \conpono{}.
We also compare against BERT-Large to contrast to a much larger language model.
We provide results from Albert \cite{lan2019albert} when available to provide a state-of-the-art baseline that may have used more data, compute and parameters.
The purpose of these results is not to compare against the current state-of-the-art but rather to better understand the improvements that can be found from adding a discourse coherence objective to BERT-Base without significantly increasing the model size or training data.

\begin{table}[h!]
\centering
\begin{tabular}{||c | c c||} 
\hline
Model & RTE & COPA \\ 
\hline\hline
BERT-Base & 66.4 & 62.0  \\ 
BERT-Base BSO & 71.1 & 67.0 \\
\conpono{} & 70.0 & 69.0  \\
BERT-Large & 70.4 & 69.0 \\
ALBERT & 86.6 & - \\
 \hline
\end{tabular}
\caption{Our model improves accuracy over BERT-Base for RTE and COPA benchmarks. Improvements are comparable to BERT-Large but still lag behind much larger models trained on more data, such as ALBERT. All scores are on the validation set.}
\label{table:glue}
\end{table}

\textbf{Results:}
We believe that the coherence and ordering aspects of these evaluation tasks are well fit to demonstrate the how our model can improve on strong baselines such as BERT-Base.
Table~\ref{table:glue} shows that our model achieves accuracies on RTE and COPA comparable to BERT-Large while having the same number of parameters as BERT-Base.
Interestingly, we observe improvements over the baseline with BERT-Base BSO, showing that even simple discourse-level objectives could lead to noticeable downstream effects.
Though these improvements are modest compared to BERT-Large, they are meant to highlight that our model does not only improve on results for artificial sentence ordering tasks, but also on aspects of benchmarks used to generally evaluate pretrained language models and language understanding.

\subsubsection{ReCoRD results and models}

\begin{table}[h!]
\centering
\begin{tabular}{||c | c  ||} 
\hline
Model & Accuracy \\ 
\hline\hline
BERT-Base &  61.2  \\ 
\conpono{} &  63.2  \\
BERT-Large &  69.8 [EM] \\
 \hline
\end{tabular}
\caption{\conpono{} is more effective at classifying the most plausible sentence from the extended context than BERT-Base.  We report the BERT-Large exact match score, where the model selects only the target entity from the context, for reference. All scores are on the validation set.}
\label{table:record}
\end{table}

\vspace{1cm}
The task for the ReCoRD dataset is to select the correct entity from those that appear in the context to fill in the blank in the target.
Previous models for ReCoRD have used a similar structure to SQuAD \cite{rajpurkar-etal-2016-squad} where the model outputs a vector for each token and the model learns the best start and end position of the answer span based on the softmax over all the tokens.
We, instead, generate all possible target sentences by filling the blank with each marked entity and discriminatively choose the sentence most likely to be the true ``plausible'' sentence from the context.
This modified task evaluates how our model compares to BERT-Base choosing the most coherent sentence from a set of nearly identical sentences.
In Table~\ref{table:record} we show that \conpono{} does achieve a boost over BERT-Base but is still well below BERT-Large exact match score on the harder task of selecting the entities in context.
The strong results from BERT-Large imply that having a better representation of the text with a large model is able to subsume any improvement from learning plausible contexts for this task.

\subsection{Ablations}\label{sec:ablation}

There are three aspects of our modeling choices that warrant a deeper understanding of their importance to the model:

\begin{itemize}
    \item \textit{Window size:} We ablate the 4 window sizes (ie. choices of k). 
    k = 1 is effectively binary sentence ordering with negative samples.
    \item \textit{Masked Language Model Objective:} We remove the MLM objective allowing the model to optimize only the \conpono{} objective without maintaining a good token level representation.
    \item \textit{Model size:} We train a smaller model that is also initialized with pretrained weights.
\end{itemize}

To measure the effects of each of these design decisions, we report DiscoEval scores for each model as well as accuracy on the \conpono{} classification task on a held-out set of examples. 
This is to show how well the model is optimized as well as how well it performs on downstream tasks.

\begin{table*}[h!]
\centering
\begin{tabular}{||c | c c c c c c c c||} 
 \hline
 Model & SP & BSO & DC & SSP & PDTB-E & PDTB-I & RST-DT & avg. \\ 
 \hline\hline
 k=4 & 59.84 & 76.05 & \textbf{73.62} & \textbf{80.65} & 42.28 & 44.25 & 63.00 & 62.81  \\
 k=3 & 60.47 & 76.68 & 72.74 & 80.30 & \textbf{43.40} & 44.28 & 62.56 & 62.92 \\
 k=2 & \textbf{60.67} & \textbf{76.75} & 72.85 & 80.38 & 42.87 & \textbf{44.87} & \textbf{63.13} & \textbf{63.07} \\ 
 k=1 & 47.56 & 66.03 & 72.62 & 80.15 & 42.79 & 43.55 & 62.31 & 59.29 \\ 
  - MLM & 54.92 & 75.37 & 68.35 & 80.2 & 41.67 & 43.88 & 61.27 & 60.81 \\
 Small & 45.41 & 61.70 & 67.71 & 75.58 & 35.26 & 36.18 & 46.58 & 52.63 \\ 
 \hline
\end{tabular}
\caption{The ablation analysis shows the effects of different $k$ values (ie. window sizes) in our objective, removing the MLM objective during pretraining and training with a small transformer encoder.}
\label{table:ablation}
\end{table*}

\begin{figure*}[h]
\includegraphics[width=16cm]{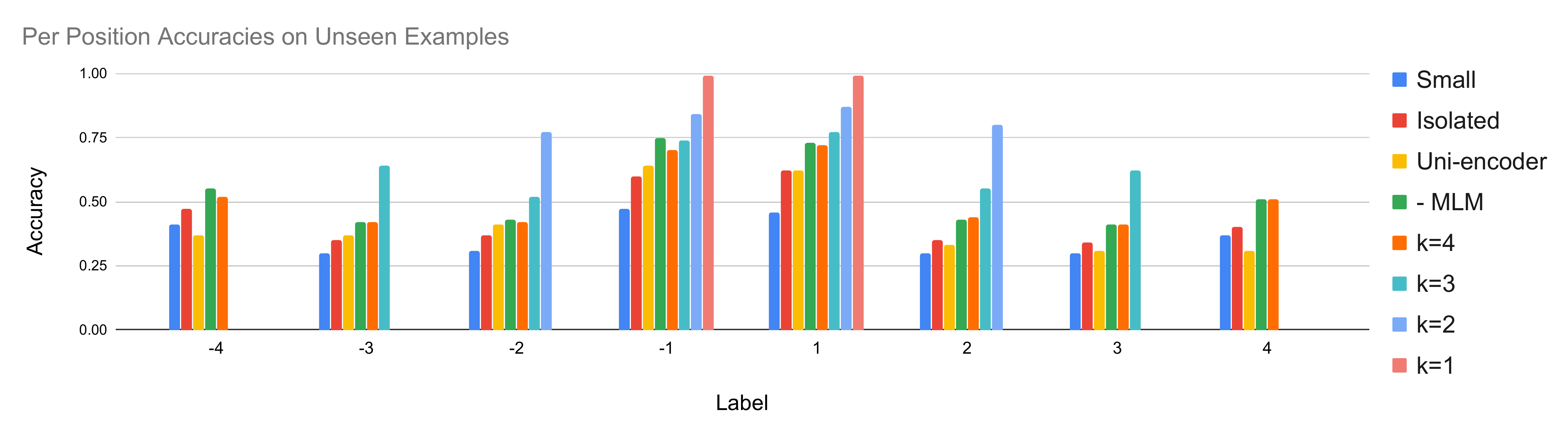}
\caption{We can evaluate the accuracy on the \conpono{} objective for each label (ie. distance between anchor and target sentence) on a set of 5,000 examples held-out from training. We observe that higher accuracy does not necessarily correlate with better downstream performance on DiscoEval.}
\label{fig:ablation}
\end{figure*}

Table~\ref{table:ablation} shows the results on DiscoEval with our model and several key ablations.
We observe that using a window size for our objective that is larger than 1 is key to seeing downstream improvements.
We believe that this is due to the objective being harder for the model because there is more variation farther from the anchor.
At the same time, increasing the window size beyond 2 seems to result in similar performance.
This may be because larger distances from the anchor also lead to more ambiguity.
We see this reflected in the held-out classification accuracy being lower for examples with larger distance labels in Figure~\ref{fig:ablation}.

We also note that keeping the masked language model objective during pretraining also improves downstream performance.
In Figure~\ref{fig:ablation} we see that classification accuracy is consistently lower with the MLM objective compared to without.
This is expected because during inference, many key terms may be masked out, making the task harder.
However, keeping this objective during pretraining maintains a good token-level representation that is necessary for downstream tasks.

Lastly, we try training a smaller version of our model, with only 2 hidden layers, and a 512 intermediate size.
The smaller model is able to train much faster, allowing us to train on many more examples and new data.
However, we are unable to achieve similar results despite training on 24 times more examples, and including CCNews \cite{liu2019roberta}, a larger and higher quality data source.

\subsection{Qualitative Analysis}

To glean some insight into how \conpono{} representations may differ from BERT-Base representations, we look at the occurrence of discourse markers in the BSO-Wikipedia task of DiscoEval.
We choose this task because it is a simple binary classification task that has only 2 sentences as input and the domain is similar to the pre-training data.
We look at the usage of discourse markers identified by \citet{nie2017dissent}; \textit{but, when, if, before, because, while, though, after, so, although, then, also, still}. 
\footnote{We omit \textit{and} and \textit{as} because they are very common in this corpus but often are not used as connectives between the two candidate sentences for the BSO task.}

We extract examples from the test set on which \conpono{} output the correct label and BERT-Base output the incorrect label and visa versa.
For each set of examples, we measure the change in the occurrence of discourse markers relative to the training data counts.
Since some markers are much more common than others, we take the weighted average of the change in appearance rate, where the weights are the training data counts of each marker.

We find that in the set of examples that \conpono{} classified correctly, the rate of discourse makers was 15\% higher than in the training corpus.
This is in contrast to 11\% higher among the examples that BERT classified correctly.
The standard deviation for random samples of the same size was about 1\%.
This suggests that both BERT and \conpono{} are relying heavily on discourse markers to solve the BSO-Wikipedia task.

While it is expected for shallow discourse markers to be strong features for sentence ordering, we expect \conpono{} to also incorporate deeper features, such as anaphora, due to its pretraining objective.
One indication of \conpono{} relying on alternative features than BERT-Base is that there was a 12\% relative increase in discourse markers in the \conpono{} set when counting markers only in the first sentence whereas an 8\% relative increase in the BERT set when counting markers only in the second sentences.
The difference in the location of the discourse markers in the two sets of examples suggests that \conpono{} and BERT utilize those features differently and that \conpono{} may be less likely to incorrectly classify examples that use discourse markers in the first sentence of a BSO example.
Manually inspecting a sample of examples hints that there are often strong coreferences between the two input sentences that indicate the ordering.



Table~\ref{table:text_examples} shows two examples from the \conpono{} correct set which is drawn from the BSO-Wikipedia test data.
In both examples, the discourse marker appears in the first sentence but the second sentence contains anaphora referring to an antecedent in the first sentence.

\begin{table*}[h!]
\centering
\small
\begin{tabular}{ p{15.5cm} } 
 \hline
 \hline
 In 1941 \underline{[1]Vaughn} joined the United States National Guard for what had been planned as a one-year assignment , \textbf{but when} \underline{[2]World War II} broke out , \underline{he} was sent abroad until the war ended in 1945 . \\
\noalign{\vskip 2mm}   
\underline{[1]He} decided to make music a career when he was discharged from the army at the end of \underline{[2]the war} , and attended Western Kentucky State College , now known as Western Kentucky University , majoring in music composition . \\
\noalign{\vskip 2mm}   
\hline
\textbf{Although} it lasted only twenty-three years ( 1933–1956 ) and enrolled fewer than 1,200 students , \underline{Black Mountain College} was one of the most fabled experimental institutions in art education and practice . \\
\noalign{\vskip 2mm}  
\underline{It} launched a remarkable number of the artists who spearheaded the avant-garde in the America of the 1960s . \\
\noalign{\vskip 2mm}   
\hline
\hline
\end{tabular}
\caption{Two examples from the DiscoEval BSO-Wikipedia test set on which \conpono{} made the correct prediction but BERT-base did not. \textbf{Bold} terms are discourse markers, \underline{underlined} terms are co-referents. In both examples, the discourse marker appears in the first sentence but the second sentence has anaphora referring to an antecedent in the first sentence.}
\label{table:text_examples}
\end{table*}

\section{Related Work}

Some of the largest improvements on benchmarks such as GLUE \cite{wang-etal-2018-glue} have come from ELMO's large scale bi-directional language modeling \cite{peters-etal-2018-deep}, BERT's masked language models \cite{devlin-etal-2019-bert}, XLNET's generalized autoregressive pretraining \cite{yang2019xlnet}, RoBERTa's robust training \cite{liu2019roberta} and ALBERT's parameter reduction techniques \cite{lan2019albert}.
As discussed in Section~\ref{sec:sent_objectives}, most language model were limited to NSP or BSO for inter-sentence representation learning.
We showed that by comparing to BERT, which uses NSP and BERT-Base BSO which we train with the BSO objective that our objective is able to improve the discourse-level representations by training on more fine-grained sentence ordering, non-contiguous neighboring sentences and contrasting against random negatives.

Early approaches to sentence representation, such as Skip-Thought Vectors \cite{kiros2015skip}, mimicked word embedding methods in addition to left-to-right language modeling to use unlabeled data to learn sentence level representations.
DisSent \cite{nie-etal-2019-dissent} focused more on collecting data that could be used to train a supervised classification model on pairs of sentences.
These and other innovations in sentence representation lead to the creation of more evaluations for discourse and coherence representation \cite{chen-etal-2019-evaluation, xu-etal-2019-cross}.

Like other unsupervised representation learning models, \conpono{} is trained to generate a latent variable that encodes inter-sentence relationship and discourse coherence.
Our objective is inspired by the Contrastive Predictive Coding (CPC) objective \cite{oord2018representation}.
CPC was originally designed to be a ``universal unsupervised learning approach to extract useful representations from high-dimensional data'' and was previously implemented on the token-level for text models.
We utilize the k-distance predictions of CPC because it naturally captures discourse and sentence ordering properties when applied on the sentence-level.
Furthermore, by combining our objective with a transformer encoder, our model is able to benefit from cross-attention between the anchor and the target sentences, which we show outperforms encoding the anchor and target separately, as implemented in CPC.
In Section~\ref{sec:ablation} we show that the cross-attention is an important factor in learning a good representation for downstream tasks and effectively optimizing our inter-sentence objective.

\section{Discussion}

In this paper we present a novel approach to encoding discourse and fine-grained sentence ordering in text with an inter-sentence objective.
We achieve a new state-of-the-art on the DiscoEval benchmark and outperform  BERT-Large with a model that has the same number of parameters as BERT-Base.
We also observe that, on DiscoEval, our model benefits the most on ordering tasks rather than discourse relation classification tasks.
In future work, we hope to better understand how a discourse model can also learn fine-grained relationship types between sentences from unlabeled data.
Our ablation analysis shows that the key architectural aspects of our model are cross attention, an auxiliary MLM objective and a window size that is two or greater.
Future work should explore the extent to which our model could further benefit from initializing with stronger models and what computational challenges may arise.



\section*{Acknowledgments}
We wish to thank the Stanford NLP group for their feedback.
We gratefully acknowledge support of the DARPA Communicating with Computers (CwC) program under ARO prime contract no. W911NF15-1-0462 


\bibliography{anthology,acl2020}
\bibliographystyle{acl_natbib}

\appendix

\section{Appendix}
We include some fine-grained DiscoEval results that were reported as averages, as well as implementation and reproduction details for our experiments.

\subsection{SP, BSO and DC breakdown}

\begin{table*}[!b]
\centering
\begin{tabular}{||c | c c c | c c c | c c||} 
 \hline
 & \multicolumn{3}{c|}{SP} & \multicolumn{3}{c|}{BSO} & \multicolumn{2}{c||}{DC}  \\
 Model & Wiki & arxiv & ROC & Wiki & arxiv & ROC & Wiki & Ubuntu \\ 
 \hline\hline
 BERT-Large & 50.7 & 47.3 & 63.4 & 70.4 & 66.8 & 70.8 & 65.1 & 54.2 \\
 RoBERTa-Base & 38.35 & 33.73 & 44.00 & 60.19 & 55.16 & 60.66 & 62.80 & 53.89 \\
 BERT-Base BSO & 49.23 & 50.92 & 60.80 & 74.67 & 68.56 & 72.22 & 88.80 & 56.41 \\
 \conpono{} - MLM & 50.95 & 51.90 & 61.92 & 77.98 & 71.45 & 76.68 & 86.70 & 50.00 \\
 \conpono{} Small & 44.90 & 41.23 & 50.10 & 65.03 & 58.89 & 61.19 & 78.10 & 57.32 \\ 
 \conpono{} \textit{isolated} & 49.33 & 44.60 & 56.53 & 59.16 & 57.48 & 56.94 & 71.60 & 54.71 \\
 \conpono{} \textit{uni-encoder} & 54.30 & 58.58 & 66.75 & 78.25 & 71.65 & 73.99 & 86.00 & 57.90 \\
 k=4 &  54.07 & 58.30 & 67.15 & 79.04 & 72.21 & 76.89 & 88.38 & \textbf{58.85} \\
 k=3 & 54.65 & \textbf{59.55} & 67.22 & \textbf{79.34} & \textbf{73.61} & \textbf{77.08} & \textbf{89.48} & 56.00 \\
 k=2 &  \textbf{54.83} & 58.77 & \textbf{68.40} & 79.24 & 74.16 & 76.84 & 89.22 & 56.41 \\
 k=1 & 44.05 & 40.98 & 57.65 & 68.47 & 62.40 & 67.24 & 89.03 & 56.20  \\ 
 \hline
\end{tabular}
\caption{SP, BSO and DC are composed of separate datasets. We report the average in the main paper but show the breakdown here.}
\label{table:finegrained}
\end{table*}

Table~\ref{table:finegrained} shows the scores for each model per each dataset domain for the SP, BSO and DC tasks in DiscoEval.

\subsection{\conpono{} pretraining details}
\conpono{} is pretrained on 1.6 million examples randomly sampled from Wikipedia and BooksCorpus.
We use the same number of training examples for all the ablations and training BERT-Base BSO.
On example consists of a single anchor and 32 candidate targets, 4 losses (1 for each of the 4 randomly chosen true targets (ie. $k$)).
We use a 25\% warm up rate and a learning rate of 5e-5.
The model is initialized with BERT-Base weights.
We add a square interaction weight matrix that is the same size as model output dimensions (ie. 756) that is referred to as $W_k$ in Section~\ref{sec:model}.
There is one such matrix for each $k$.
The maximum sequence length of the input is 512, though do to some preprocessing constraints, the maximum input seen by the model is 493.

Our \conpono{} small model has a hidden size of 128, an intermediate size 512, and has 2 hidden layers.
We train it on 38.4 million examples, including examples from CCNews.
Samples are drawn from each source proportional to the size of the source, meaning that about 70\% of training examples come from CCNews.
Otherwise, we use all the same parameters as \conpono{}.

\subsection{Parameter counts}

Table~\ref{table:param_counts} shows the number of parameters in each model used.

\begin{table}[h!]
\centering
\begin{tabular}{||c | c  ||} 
\hline
Model & Parameters \\ 
\hline\hline
BERT-Base &   110M  \\ 
RoBERTa-Base & 110M \\
\conpono{} [All Variants] &   110M  \\
BERT-Large &  335M  \\
 \hline
\end{tabular}
\caption{}
\label{table:param_counts}
\end{table}

\subsection{RTE, COPA and ReCoRD details}

\textbf{RTE} is trained for 3240 steps, with checkpoints every 750 steps and a learning rate of 8e-6.
The warm-up proportion is 10\% and the a maximum sequence length of 512

\textbf{COPA} is trained for 300 steps, with checkpoints every 50 steps and a learning rate of 1e-5.
The warm-up proportion is 10\% and the maximum sequence length of 512.

\textbf{ReCoRD} is trained for 8 epochs over the training data with a learning rate of 2e-5, warm-up proportion of 10\% and a maximum sequence length of 512.

\end{document}